\documentclass{article}

\usepackage[final]{neurips_2022}

\usepackage[utf8]{inputenc} 
\usepackage[T1]{fontenc}    
\usepackage{hyperref}       
\usepackage{url}            
\usepackage{booktabs}       
\usepackage{multirow}
\usepackage{tabulary}
\usepackage{amsfonts}       
\usepackage{nicefrac}       
\usepackage{microtype}      
\usepackage{xcolor}         
\usepackage[framemethod=tikz]{mdframed} 

\usepackage{amsmath}
\usepackage{amsfonts}
\usepackage{graphicx}

\title{Scaling Laws Beyond Backpropagation}

\author{Matthew J. Filipovich$^{1, 2}$ \hspace{0.4cm} Alessandro Cappelli$^{1}$ \hspace{0.4cm} Daniel Hesslow$^{1}$ \hspace{0.4cm} Julien Launay$^{1, 3}$ \vspace{0.2
cm} \\
 $^1$LightOn \hspace{0.4cm} $^2$Queen's University  \hspace{0.4cm} $^3$LPENS, École Normale Supérieure \vspace{0.2cm}\\
  \texttt{\{firstname\}@lighton.ai}
}

\begin{document}

\maketitle

\begin{abstract}
Alternatives to backpropagation have long been studied to better understand how biological brains may learn. Recently, they have also garnered interest as a way to train neural networks more efficiently. By relaxing constraints inherent to backpropagation (e.g., symmetric feedforward and feedback weights, sequential updates), these methods enable promising prospects, such as local learning. However, the tradeoffs between different methods in terms of final task performance, convergence speed, and ultimately compute and data requirements are rarely outlined. In this work, we use scaling laws to study the ability of Direct Feedback Alignment~(DFA) to train causal decoder-only Transformers efficiently. Scaling laws provide an overview of the tradeoffs implied by a modeling decision, up to extrapolating how it might transfer to increasingly large models. We find that DFA fails to offer more efficient scaling than backpropagation: there is never a regime for which the degradation in loss incurred by using DFA is worth the potential reduction in compute budget. Our finding comes at variance with previous beliefs in the alternative training methods community, and highlights the need for holistic empirical approaches to better understand modeling decisions.
\end{abstract}

\section{Introduction}

Backpropagation (BP) \cite{Werbos:74, Rumelhart:86} is just one of many ways to solve the credit assignment problem--and hence to train neural networks. BP estimates the  individual contribution of each parameter to the error, but approximate methods can be employed: either through fundamentally different approaches (e.g. Hebbian learning) \cite{ movellan1991contrastive, o1996biologically}, or by relaxing constraints of BP \cite{lillicrap2016random, frenkel2021learning, jaderberg2017decoupled}. Beyond backpropagation methods have a history of being studied to understand how biological brains may learn \cite{crick1989recent, lillicrap2020backpropagation}. Indeed, key features of backpropagation are not possible to implement under biological constraints. For instance, using the transpose of the weights $\mathbf{W}^T$ in the feedback path is not possible, as the feedforward and feedback pathways are physically distinct--this is known as the weight transport problem \cite{grossberg1987competitive}. 

Once relegated to toy problems \cite{bartunov2018assessing}, alternatives to BP have now been demonstrated to be able to achieve competitive performance on challenging tasks across a variety of architectures \cite{dfa_scales}. This could result in more efficient training: for instance, local learning  may enable easier parallelization of computations at scale \cite{jaderberg2017decoupled, nokland2019training}. Alternative methods may even be co-designed with hardware \cite{launay2020hardware, frenkel2020bottom}: either for novel systems such as photonic co-processors \cite{launay2020light} and memristors \cite{ernoult2019using}, or to circumvent distributed communication bottlenecks in the large clusters used to train state-of-the-art models \cite{laskin2020parallel}.

However, the tradeoffs between BP and alternatives are not always clear. If a method enables 25\% faster training, is it worth a 5\% decrease in end-task performance? Or would a model trained with BP using a 25\% smaller compute budget still be better? As most works usually only offer a few cherry-picked datapoints, this is a difficult question to answer. Instead, deriving scaling laws~\cite{kaplan2020scaling} may provide a more complete picture: by obtaining the full power-law relationship between compute spent and task performance, one may easily identify regimes in which the alternative method is competitive with BP, and even  extrapolate results to larger scales. 

\paragraph{Contributions.} We use scaling laws to study an alternative to BP, with the following contributions:
\begin{itemize}
    \item Drawing inspiration from work using scaling laws to evaluate modeling decisions~\cite{ghorbani2021scaling, chatelain2022number, bansal2022data}, we are the first to \textbf{use scaling laws to study an alternative to backpropagation}.
    \item At variance with previous beliefs \cite{dfa_scales}, \textbf{we find that the gains in compute efficiency from using the alternative method studied are never worth the degradation in performance}. This holds even if we consider the use of exotic hardware, such as optical co-processors~\cite{brossollet2021lighton}, which would offload some computations and effectively make the alternative method "free".
\end{itemize}

\section{Framing}
\paragraph{Can alternative training methods accelerate neural network training?} Surveying the current state-of-the-art, one may find numerous claims of alternative training methods achieving \textit{competitive} performance with BP across a variety of settings and tasks (e.g., \cite{belilovsky2019greedy, dfa_scales, laskin2020parallel}). 

We seek to study this claim, with three restrictions in scope: 

\begin{enumerate}
    \item We focus on Direct Feedback Alignment
 \cite{nokland2016direct}, due its simplicity and wide applicability \cite{dfa_scales}, as well as its broad hardware prospects \cite{launay2020hardware, frenkel20180, filipovich2021monolithic}, and theoretical background \cite{refinetti2021align}.
    \item We study compute-efficiency specifically (i.e, best performance achievable for a given compute budget), as this usually a significant bottleneck for scaling-up models.
    \item We conduct our study on "GPT-like" \cite{brown2020language} causal decoder-only Transformers trained on English data. These models are known to possess smooth scaling laws \cite{kaplan2020scaling, henighan2020scaling}. Because of their unique abilities \cite{Wang2022WhatLM}, they also command some of the largest training budgets in machine learning \cite{chowdhery2022palm}, making them a prime target for more compute-efficient training. 
\end{enumerate}
    
These restrictions lead us to test the following hypothesis: 

\begin{mdframed}
\textbf{Hypothesis.} Direct Feedback Alignment can train causal decoder-only models more efficiently than backpropagation, achieving better performance for a given compute budget. 
\end{mdframed}

\paragraph{Scaling laws as a holistic empirical tool.} Scaling laws have been proposed as an empirical approach to connect hyperparameters of neural networks (e.g., parameter count, training dataset size) to their performance. They have been derived both on specific downstream tasks \cite{hestness2017deep, alabdulmohsin2022revisiting} and on upstream modeling loss \cite{kaplan2020scaling}. Scaling laws can characterize the influence of data \& modeling decisions \cite{chatelain2022number, bansal2022data}, or even unveil new, more optimal training practices \cite{hoffmann2022training, sorscher2022beyond}. 

\begin{figure}[b]
    \centering
    \includegraphics[width=0.65\textwidth]{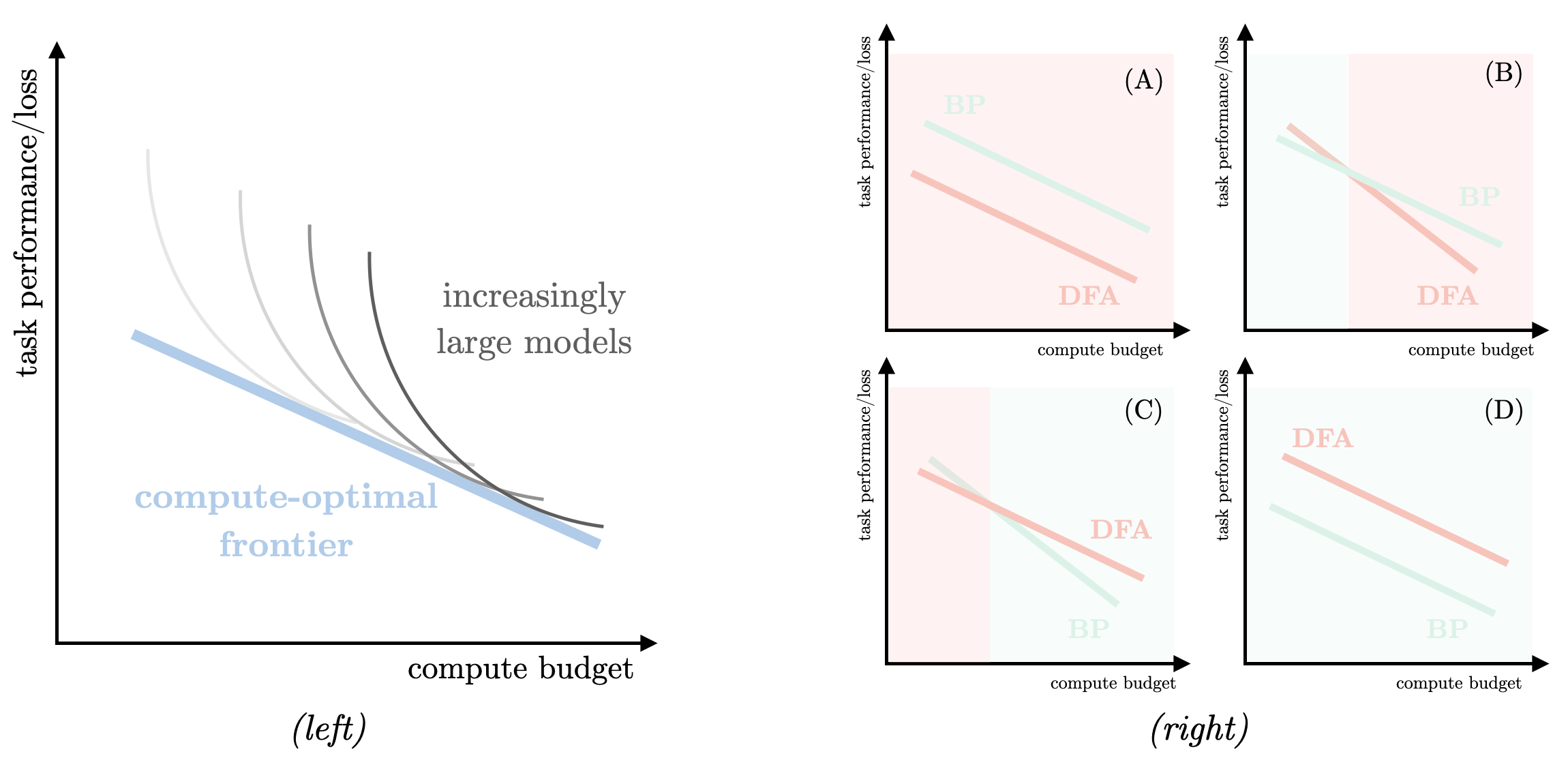}
    \caption{\textbf{Scaling laws provide optimal compute frontiers}. Left: compute-optimal frontier. Right: scenarios for DFA (red) \& BP (green) scaling laws, shading is best method at a compute budget.}
    \label{fig:scaling_crash_course}
\end{figure}

As illustrated in Figure \ref{fig:scaling_crash_course}, it is possible to derive a so-called \emph{compute optimal frontier} for a class of models: this defines $\mathcal{L}(C)$, the best performance $\mathcal{L}$ achievable for a compute budget~$C$. We fit a power-law $\mathcal{L}(C) = (C_c C)^{\alpha_C}$ over the Pareto front of multiple runs, as proposed in \cite{kaplan2020scaling}. $C_c$ is a constant offsetting the frontier, while $\alpha_C$ controls the slope. Improvements in $\alpha_C$ are rare~\cite{ghorbani2021scaling, bansal2022data}, but valuable as they would point to modeling decisions leading to increased gains at scale.

\paragraph{Potential outcomes.} We identify four potential outcomes for our study, illustrated in Figure \ref{fig:scaling_crash_course}:
\begin{itemize}
    \item[(A)] \textbf{DFA always outperform BP.} Thanks to a better offset or improvements in scaling, the increased compute-efficiency from DFA makes it always favorable to BP.
    \item[(B)] \textbf{DFA outperforms BP at scale.} Thanks to improvements in scaling (e.g., increased efficiency compared to BP at scale), DFA eventually makes the training of large models more efficient.
    \item[(C)] \textbf{BP outperforms at scale.} DFA may exhibit poor scaling behavior \cite{bartunov2018assessing}, and may not scale to larger models, leading to BP eventually outperforming DFA.
    \item[(D)] \textbf{BP always outperform DFA.} The degradation in performance observed with DFA may never be worth the potential gains in compute-efficiency.
\end{itemize}

Both (A) and (B) may be viewed as validating our hypothesis, as they both potentially motivate the use of DFA over BP. (C) and (D) would however be negative outcomes, either restraining the efficient applicability of DFA to small models, or indicating that DFA is never competitive with BP.

\section{Methods}

\paragraph{Direct Feedback Alignment.} Direct Feedback Alignment (DFA) \cite{nokland2016direct} is an extension of Feedback Alignment (FA) \cite{lillicrap2016random} which uses a random projection of the global error to directly train each layer.

We introduce at layer $i$: $\mathbf{W}_i$ the weights, $\mathbf{a}_i$ the pre-activations, $f$ the non-linearity and its derivative $f'$, $\mathbf{h}_i$ the activations, $\delta x$ the derivative of the loss against $x$, and $\odot$ the Hadamard product. DFA replaces the backpropagated signal from the $(i+1)$-th layer $\mathbf{W}_{i+1}^{T}\delta \mathbf{a}_{i + 1}$ by a random projection of the global error~$\mathbf{Be}$. For most common losses, this error $\mathbf{e}$ is simply the difference between targets and predictions. Accordingly, the update at layer $i$ is now  $\delta \mathbf{W}_i = - [(\mathbf{B} \mathbf{e}) \odot f'(\mathbf{a}_i)] \mathbf{h}_{i-1}^T$. $\mathbf{B}$, the fixed random Gaussian matrix, can be shared across all layers \cite{launay2019principled}, reducing memory and compute costs significantly--as a single  $\mathbf{Be}$ is now calculated and used for all layers . With DFA, the update now does not depend on the backward of other layers; thus, once the forward pass is complete, all of the layers can be updated concurrently, achieving so-called backward-unlocking \cite{jaderberg2017decoupled}.

Learning with DFA is made possible through a process called alignment. During the training, the forward weights will eventually align with the fixed backward weights, enabling updates which approximate backpropagation \cite{refinetti2021align}. This is best illustrated in the simpler case of FA \cite{lillicrap2016random}. For FA, the learning signal still comes from the (i+1)-th layer: $\mathbf{B}_{i+1}\delta \mathbf{a}_{i + 1}$. For this to approximate BP, we only need $\mathbf{W}_{i+1}^{T} \sim \mathbf{B}_{i+1}$. Altough we don't report on it in this work, this is a valuable diagnostic tool when experimenting: at any step, it is possible to measure the angle (cosine similarity) between the gradients predicated by backpropagation and the ones approximated by DFA. Higher alignment values are usually correlated with networks which achieve better end-task performance \cite{launay2019principled, dfa_scales}. 

\paragraph{Scaling laws for compute-efficiency.} We are interested in scaling according to compute budget, $\mathcal{L}(C)$. We split $C = C_\text{F} + C_\text{B} + C_\text{U}$, for computing the forward pass, backpropagation of the error, and weight updates respectively. For causal decoder-only models, each phase costs roughly $2ND$ (in FLOP) with $N$ the number of model parameters and $D$ the dataset size in tokens~\cite{kaplan2020scaling}--the factor 2 coming from the multiply-accumulate. Hence, $C^\text{BP} = 6ND$. When using DFA, the backpropagation of the error is not necessary, and instead a single random projection $\mathbf{Be}$ is shared across all layers. Accordingly, $C^\text{DFA}_\text{B} = 2 d_\text{model}^2 D$, as $\mathbf{B}$ is of shape  $(d_\text{model}, d_\text{model})$. Because $N \simeq 12 n_\text{layer} d_\text{model}^2$, $C^\text{DFA}_\text{B} \ll C^\text{BP}_\text{B}$. We will neglect it and consider $C^\text{DFA} = 4ND$, a $\sim 30\%$ improvement.

Finally, note that this only takes into account improvements in the FLOP compute budget. However, practitioners are usually constrained by the actual compute budget in dollars, which is best represented by the number of GPU-hours required. $C_\text{FLOP}$ and $C_\text{GPU-hours}$ can be linked through the throughput $T$ achieved per GPU, in TFLOPS. Alternative training methods like DFA may improve this throughput, by enabling increased parallelization and reducing communication bottlenecks. Nevertheless, state-of-the-art methods already achieve hardware utilization of $\sim 50\%$ at scale \cite{korthikanti2022reducing}: at best, a 2x improvement can be expected. We thus also introduce $\tilde{C}^\text{DFA} = 2ND$, a (very) optimistic estimation which supposes DFA would enable a doubling in effective throughput. In practice, current implementations of DFA are not optimized, and it is unrealistic for DFA to be able to lift all bottlenecks currently encountered in distributed training--we use this estimate as an absolute lower bound of what is possible.

\begin{figure}[t]
    \centering
    \includegraphics[width=\textwidth]{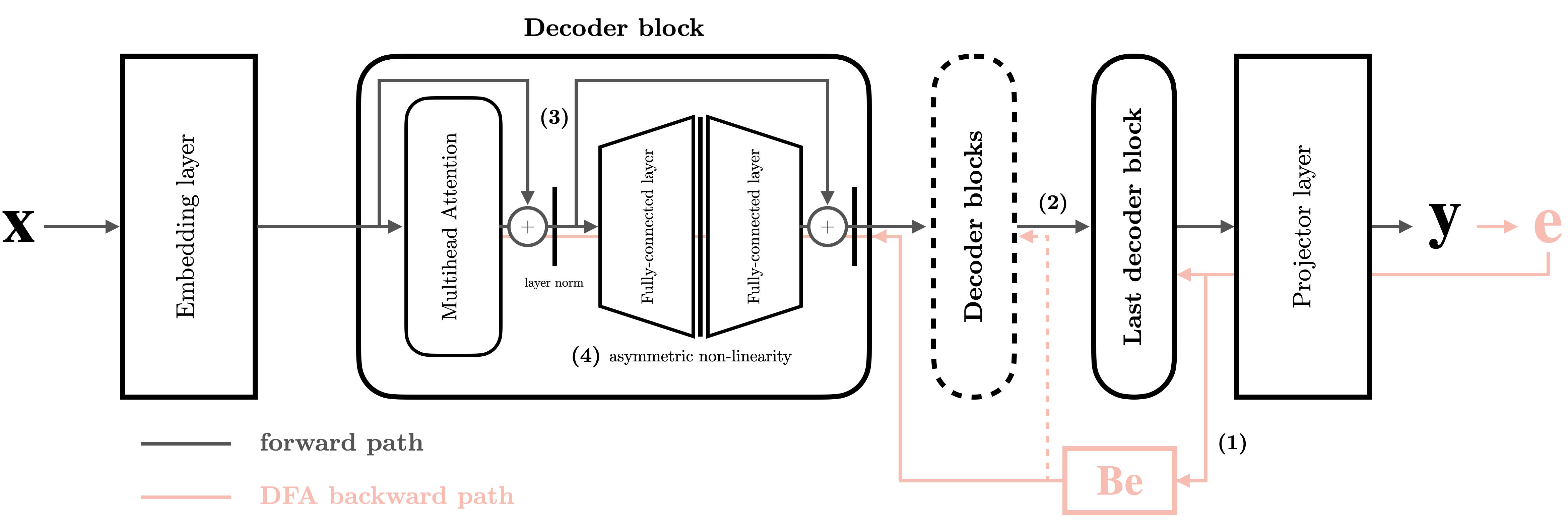}
    \caption{\textbf{Our implementation of DFA in decoder-only models}.  (1): we take the error before the projector to reduce its dimensionality; (2) we apply DFA per block and backpropagate the DFA signal within a decoder block; (3) residuals are only used during the forward and ignored in the backward; (4) we use ReLU as a non-linearity during the forward, but use the derivative of $\tanh$ in the backward instead. Note that the last decoder block is trained in the same way it would with BP}
    \label{fig:dfa_decoder}
\end{figure}

\paragraph{Training Transformers with DFA.} We train decoder-only Transformer models with both DFA and BP on a causal language modeling objective \cite{radford2018improving, radfordlanguage}, scaling from 60M to 500M parameters. The models are trained on 30B tokens of English CommonCrawl data filtered with CCNet \cite{wenzek2020ccnet}, with hyper-parameters adapted from GPT-3 \cite{brown2020language}, but with a context size of 512 for increased training speed. We chose different optimization hyper-parameters for DFA and BP, performing a sweep for each run and using the best set found for each method. We also train a so-called shallow baseline, as recommended by \cite{launay2019principled}. This shallow approach only trains the topmost layer, and provides a baseline if DFA was not training deeper layers at all. Since the scaling law for the shallow baseline is easily predictable even at small scale we do not train the largest model in order to save compute.

To improve performance, we diverge from the canonical implementation of DFA (see Figure \ref{fig:dfa_decoder}, or Table \ref{tab:overview} in appendix for a comparison between all training methods considered):
\begin{itemize}
    \item[(1)] \textbf{Preprojector error.} We take the error $\mathbf{e}$ immediately before the final projector layer instead of after. This reduces the dimension of $\mathbf{e}$ from the size of the vocabulary (51,200) to $d_\text{model}$ (576-1,408 for our runs). We found that not only does this decrease memory needs and compute costs, but this also results in a small improvement in autoregressive loss.
    \item[(2)] \textbf{Block-wise DFA.} We only apply DFA per decoder-block, and then backpropagate the DFA signal within the block, significantly improving autoregressive loss \cite{dfa_scales}. This maintains the ability to update blocks independently from one another, simplifying classic parallelization schemes such as pipeline parallelism.  However, this makes $C_\text{B}^\text{DFA}$ non-negligible. In our plots, we still neglect $C_\text{B}^\text{DFA}$, making our comparisons strongly biased toward DFA. As BP can leverage decades of research and methods finetuned for its idiosyncrasies, we believe this is a fair way to offer a best case scenario for DFA.
    \item[(3)] \textbf{Asymmetric residuals.} We make the residual paths asymmetric: although they remain untouched in the feedforward, they are disabled and ignored in the backward. Within blocks, this prevent the DFA feedback from directly flowing through the attention--\cite{dfa_scales} found that attention layers often struggled to align when a DFA feedback was applied directly. Letting the DFA feedback first flow through the fully-connected layers improves alignment. 
    \item[(4)] \textbf{Asymmetric non-linearities.} Previous work identified that DFA performs best with activation functions which are continuous and bounded \cite{launay2019principled}; the classic activation functions used in Transformers, ReLU and GeLU, do not fit this criteria. Switching to $\tanh$ reduced the gap between DFA and BP, but also worsened autoregressive loss overall. Instead, we keep ReLU in the forward, but replace its derivative with the derivative of $\tanh$ in the backward. This asymmetric approach improves both alignment and loss.
\end{itemize} 

Finally, note our experiments use 16 to 32 A100 80GB GPUs, using data parallelism only.  

\section{Experiments}

\begin{figure}[t]
    \centering
    \includegraphics{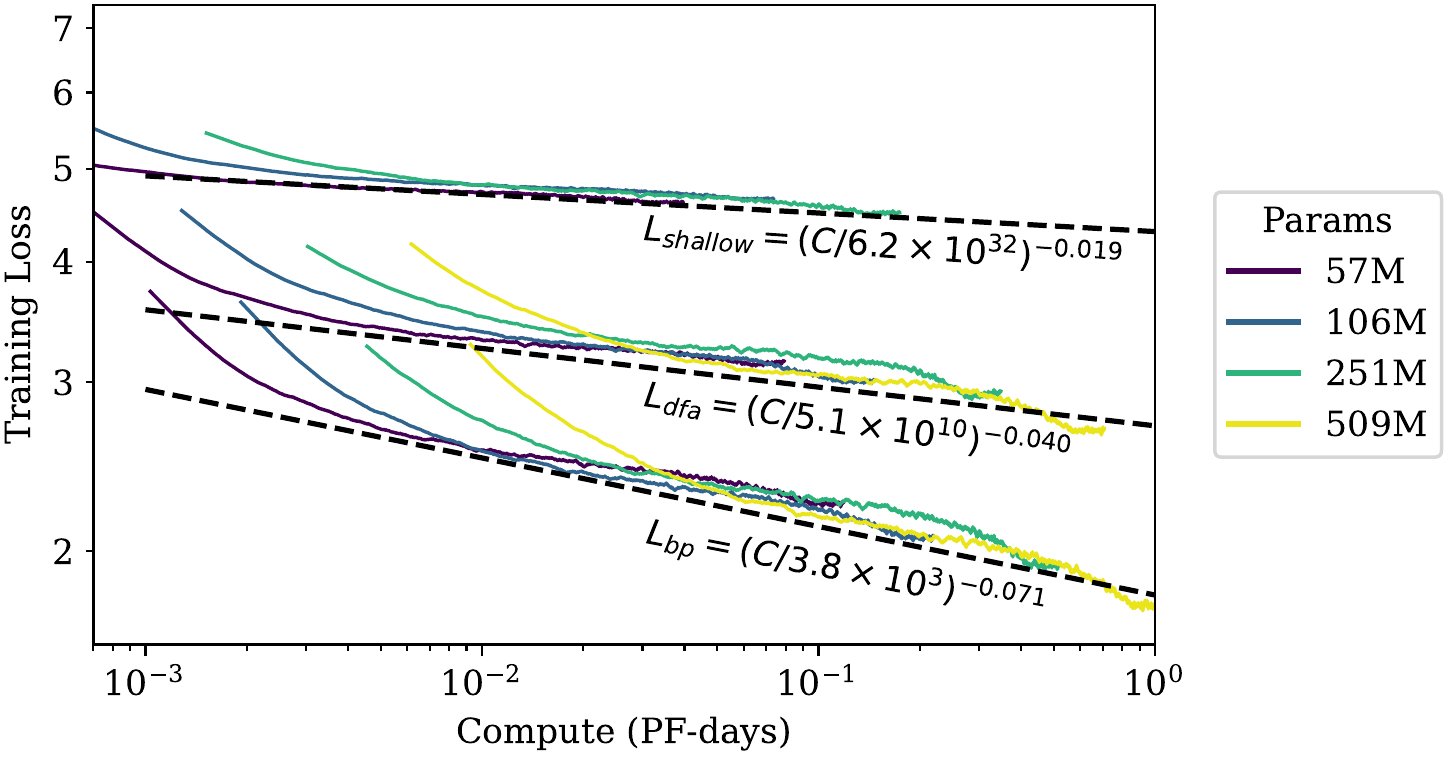}
    \caption{\textbf{The degradation in loss incurred by DFA is never worth the improvement in compute.} We plot the training loss against the compute expended--which is $\sim 30\%$ smaller for DFA thanks to improved efficiency. The compute-optimal frontier of DFA is significantly worse than that of BP, and no practical model would benefit from using DFA instead of BP, falsifying our original hypothesis.}
    \label{fig:scaling_laws}
\end{figure}

We train models over an order of magnitude in scale: from 57 million to 509 million parameters. These are the largest models ever trained with DFA, representing a significant departure from usual small-scale experiments in the alternative training methods community. 

\paragraph{Results.} Our results are showcased in Figure \ref{fig:scaling_laws}, with best fit parameters for the compute-optimal frontier in Table \ref{tab:best_fit}. We report the training loss directly, as we perform a single epoch over the 30B tokens of data. For the shallow baseline, $C=2ND$ as we neglect the cost of updating the topmost layer only. Compute budgets are in PF-days, one PFLOPS of compute power sustained for a day.

\begin{table}[b]
\centering
\caption{\textbf{Best power-law fit of the compute-optimal frontier}. We fit $\mathcal{L}(C) = (C_c C)^{\alpha_C}$ to the curves of Figure \ref{fig:scaling_laws}, lower values of $\alpha_C$ denote better scaling and lower values of $C_c$ an improved offset. Although DFA updates do train layers (the performance observed is better than the shallow baseline), scaling is significantly degraded compared to backpropagation.}
\label{tab:best_fit}
\begin{tabular}{@{}lcc@{}}
\toprule
 & $\alpha_C$      & $C_c$ [$\text{PF-days}^{-1}$]                       \\ 
\midrule
\textbf{BP}              & \underline{-0.071} & \underline{$3.8 \times 10^3$} \\
\textbf{DFA}             & -0.040          & $5.1 \times 10^{10}$       \\
\textbf{Shallow}         & -0.019          & $6.2 \times 10^{32}$       \\ \bottomrule
\end{tabular}
\end{table}

First, we note that DFA performs significantly above the shallow baseline. This confirms that layers trained by DFA are actually learning meaningful representations, rather than remaining random throughout training. The shallow baseline exhibit little to no scaling, limited to the increase in width of the decoder-block as we increase overall parameter count. Random parameters from additional untrained layers are unlikely to contribute to scaling, as reported by \cite{chatelain2022number}.

Importantly, there is never a region for which using DFA is preferrable to using backpropagation from a compute-effiency perspective. DFA not only presents a worse offset than BP, it scales more poorly: the gap with BP widens as compute budget and model scale increases. This is scenario (D): BP always outperform DFA, thus falsifying our original hypothesis.

\section{Discussion}

\paragraph{Limitations.} As for all negative results, absence of evidence is not evidence of absence. Novel modeling practices may enable DFA to better train decoder-only models, bridging the gap with BP. We did however explore and implement many of such ideas (e.g., preprojector error, asymmetric residuals/non-linearities), failing to bridge said gap on our own. Our study is also focused on the specific setting of causal decoder-only models, for which scaling laws are easy to derive. Such large language models are well-known to be difficult to train, even with BP \cite{zhang2022opt}. DFA may find more success with simpler architectures. Furthermore, our study did not consider other alternatives to backpropagation: like for DFA, promising results exist also for greedy layerwise methods \cite{belilovsky2019greedy} or local gradients approaches \cite{jaderberg2017decoupled}. Finally, we did not conduct a full exploration of learning rate schedules and architectural hyperparameters, as performed in recent scaling laws studies \cite{hoffmann2022training} to find the compute-optimal frontier. However, given the large gap between BP and DFA, it is unlikely such an exploration would result in a significantly  different finding.

\paragraph{Conclusion.} Alternative training methods exhibit compelling properties: they may enable local learning \cite{nokland2019training}, leading to easier and better parallelization of distributed computations \cite{laskin2020parallel}, and may even open new avenues for exotic hardware \cite{dabos2022neuromorphic, launay2020hardware}. However, care must be taken to make fair comparisons. Notably, reporting individual datapoints can be misleading, and more holistic approaches, such as scaling laws, should be favored to paint the full picture of tradeoffs. 

In this work, we used scaling laws to characterize the compute-efficient frontier of DFA when training causal decoder-only models, and compared with backpropagation. We originally hypothesized, following previous literature, that the increase in compute efficiency from DFA may lead to a favorable compromise against backpropagation. However, our experiments falsified this hypothesis:
\begin{mdframed}
    \textbf{Finding.} At variance with previous beliefs, using Direct Feedback Alignment to train causal decoder-only models is never more compute-efficient than using backpropagation.
\end{mdframed}

This finding holds despite assumptions heavily in favor of DFA: even if we were to assume a compute budget of $\tilde{C}^\text{DFA} = 2ND$ (removing entirely the cost of the backward and updates, for instance by offloading them to a coprocessor \cite{launay2020light, frenkel2021learning}), DFA is never more compute-efficient than backpropagation.

Although this is a significantly negative finding, we would like to end on two outlooks:
\begin{itemize}
    \item \textbf{Alternative methods beyond compute-efficiency.} Our work studied DFA under the practical light of compute-efficiency, but alternative training methods can also be motivated by more than simply reducing compute budgets. For instance, they have been invaluable in providing models for how the brain may learn \cite{lillicrap2020backpropagation}. Specifically in the case of DFA, they can also be leveraged for increased adversarial robustness \cite{cappelli2021ropust, sanfiz2021benchmarking, cappelli2022adversarial}, or for novel implementations of differential privacy \cite{ohana2021photonic, lee2020differentially, asadian2022self}. These prospects naturally encourage further work in this direction, beyond simple compute-efficiency.
    \item \textbf{Scaling laws as a key modeling tool.} The pitfalls outlined by our study also apply to other modeling works. In the specific case of Transformer models, identifying best practices has been the subject of much contention in the literature~\cite{narang2021transformer, le2022language}. Scaling laws have been invaluable in going beyond cherry-picked datapoints: for neural machine translation \cite{ghorbani2021scaling}, mixture-of-experts models \cite{clark2022unified}, or even broad architectural choices~\cite{tay2022scaling}. 
\end{itemize}

\newpage


\bibliographystyle{unsrt}
\bibliography{bibliography}

\newpage
\appendix

\section{Overview of differences between training methods}

\begin{table}[h]
\centering
\caption{\textbf{Overview of the training methods considered.}. We make a number of changes to the canonical implementation of DFA to enhance its ability to train Transformer models.}
\label{tab:overview}
\scriptsize
\begin{tabular}{p{1.6cm}p{3.7cm}p{2cm}p{2cm}p{2cm}}
\toprule
                                  & \textbf{BP}                                                                                    & \textbf{DFA}                                                      & \textbf{}                                                                                                & \textbf{Shallow}                                            \\
                                  &                                                                                                & Canonical                                                         & Ours                                                                                                     &                                                             \\ \midrule

\textbf{Training cost}                & $6ND$                                                                                          & $4ND$                                                             & $\sim 6ND$                                                                                               & $\sim 2ND$                                                  \\ 
\textbf{Update rule} & $- [(\mathbf{W}_{i+1}^{T} \delta \mathbf{a}_{i+1}) \odot f'(\mathbf{a}_i)] \mathbf{h}_{i-1}^T$ & \multicolumn{2}{l}{$- [(\mathbf{B} \mathbf{e}) \odot f'(\mathbf{a}_i)] \mathbf{h}_{i-1}^T$}                                                                                  & no update                                                   \\
\textbf{Error}                & N/A & Directly from loss & From gradient of loss before projector & N/A                                                                                                                                                                                                                                                                                                         \\

\textbf{Strategy}                 & Full backpropagation                                                              & DFA feedback at every layer, no backpropagation & Block-wise: DFA feedback at the top of decoder blocks, backpropagation within the blocks & All decoder blocks but the last one frozen \\ 
\textbf{Last block}                & \multicolumn{4}{c}{Trained with backpropagation}                                                                                                                                                                                                                                                                                                         \\
\textbf{Residuals}             & Vanilla                                                                                             & Vanilla                                                                & Asymmetric                                                                                                      & Vanilla                                                          \\
\textbf{Non-linearity}      & Vanilla                                                                                             & Vanilla                                                                & Asymmetric                                                                                                      & Vanilla      \\
\bottomrule
\end{tabular}
\end{table}

\end{document}